\documentclass{ifacconf}

\usepackage{graphicx}      
\usepackage{natbib}        
\usepackage{amsmath,amssymb,stmaryrd,mathtools, amsfonts}
\usepackage{bbm}
\usepackage{hhline}
\usepackage{multirow}%
\usepackage{booktabs}
\usepackage{enumitem}
\usepackage{url}

\usepackage{subcaption}
\usepackage[
  tableposition = top, 
  width = 0.85\textwidth 
]{caption}
\begin{document}
\begin{frontmatter}

\title{Multi-vehicle Conflict Resolution in Highly Constrained Spaces by Merging Optimal Control and Reinforcement Learning}


\author{Xu Shen, Francesco Borrelli}

\address{Department of Mechanical Engineering, University of California at Berkeley, Berkeley, CA 94701 USA (e-mail: xu\_shen@berkeley.edu)}

\begin{abstract}                
We present a novel method to address the problem of multi-vehicle conflict resolution in highly constrained spaces. An optimal control problem is formulated to incorporate nonlinear, non-holonomic vehicle dynamics and exact collision avoidance constraints. A solution to the problem can be obtained by first learning configuration strategies with reinforcement learning (RL) in a simplified discrete environment, and then using these strategies to shape the constraint space of the original problem. Simulation results show that our method can explore efficient actions to resolve conflicts in confined space and generate dexterous maneuvers that are both collision-free and kinematically feasible.
\end{abstract}


\end{frontmatter}

\section{Introduction}
\vspace{-0.1em}
Current autonomous vehicles (AVs) operate reasonably well in environments where traffic rules are well-defined, the surrounding agents are rational, and their actions can be easily predicted. However, AVs are often designed conservatively to yield for safety in environments that involve complex interactive scenarios, such as the one in Fig.~\ref{fig:scenario}. This leads to deadlocks and congestion.
Extensive research has been conducted to resolve the conflicts with machine intelligence in such scenarios. A widely adopted framework is to optimize the longitudinal motion of vehicles along their pre-defined routes. \cite{campos_cooperative_2014} solved the coordination problem with a pre-defined decision order heuristics, while \cite{murgovski_convex_2015} tried to optimize over all permutations of crossing sequences by convexifying the safety constraints, which are used by \cite{riegger_centralized_2016} to formulate a centralized Model Predictive Controller (MPC) for optimal control. \cite{katriniok_distributed_2017} designed a distributed MPC with constraint prioritization and \cite{rey_fully_2018} fully decentralized the control with the alternating direction method of multipliers (ADMM).

An alternative to formulating conflict resolution as an optimization problem, is explored by the line of research that uses reinforcement learning (RL) to learn control policies through extensive simulations.
\cite{li_game_2018} modeled the conflict resolution problem as a Markov Decision Process (MDP), and optimal actions are solved by explicitly maximizing the expected cumulative reward in a level-$k$ game setting. \cite{li_optimizing_2019} used deep RL to choose optimal actions with the state-action value function approximation. \cite{yuan_deep_2022} further proposed an end-to-end decision-making paradigm with raw sensor data, which added a Long Short-Term Memory network into deep RL. Among these research, the MDP formulation often requires discrete action space for efficient learning.

When conflicts arise in highly constrained spaces such as crowded parking lots, both the optimal control and the RL approaches often fail due to the following reasons:
\begin{enumerate}[label=(\roman*)]
    \item The vehicles need to plan for combinatorial actions in order to create spaces for each other to pass through;
    \item The nonlinear and non-holonomic vehicle dynamics cannot be neglected since the vehicles are sensitive to model errors in close proximity to obstacles;
    \item The environment is non-convex and geometrical approximation can easily run out of free space;
    \item The vehicles need to perform complex maneuvers to navigate in this environment, which often exploits their full motion capacities, such as frequent gear switching and steering saturation.
\end{enumerate}

Solving such a highly nonlinear and non-convex optimal control problem is intractable without proper reformulation and good initial guesses. In this work, we propose a novel method to solve such a class of problems by merging RL and optimal control, which can generate dexterous maneuvers for multiple vehicles to resolve conflicts in tightly-constrained spaces.
Our contributions are:
\begin{enumerate}[label=(\roman*)]
    \item We model the conflict resolution problem as a multi-agent partially observable Markov decision process while preserving vehicle body geometry and non-holonomic dynamics. 
    A shared deep Q-network (DQN) policy is trained to drive all agents towards their destinations safely and efficiently.
    \item We leverage the trained policy to generate 
    strategies to guide the vehicle configurations in the optimal control problem.
    To obtain solutions to this non-convex, nonlinear programming (NLP) problem, we also provide an approach to compute effective initial guesses. The proposed approach uses a sequence of steps starting from the DQN policy output and progressively refining with simple NLP problems. 
\end{enumerate}

In the remainder of this paper: Section~\ref{sec:formulation} formulates conflict resolution as an optimal control problem. Section~\ref{sec:rl} uses RL to search for strategies of tactical vehicle configurations. Section~\ref{sec:planning} leverages these strategies with progressive steps to obtain optimal solutions. We present the numerical details and simulation results in Section~\ref{sec:results}.
\section{Problem Formulation}
\label{sec:formulation}

\subsection{Example Scenario}
\label{sec:scenario}

\begin{figure}
\begin{center}
\includegraphics[width=0.9\linewidth]{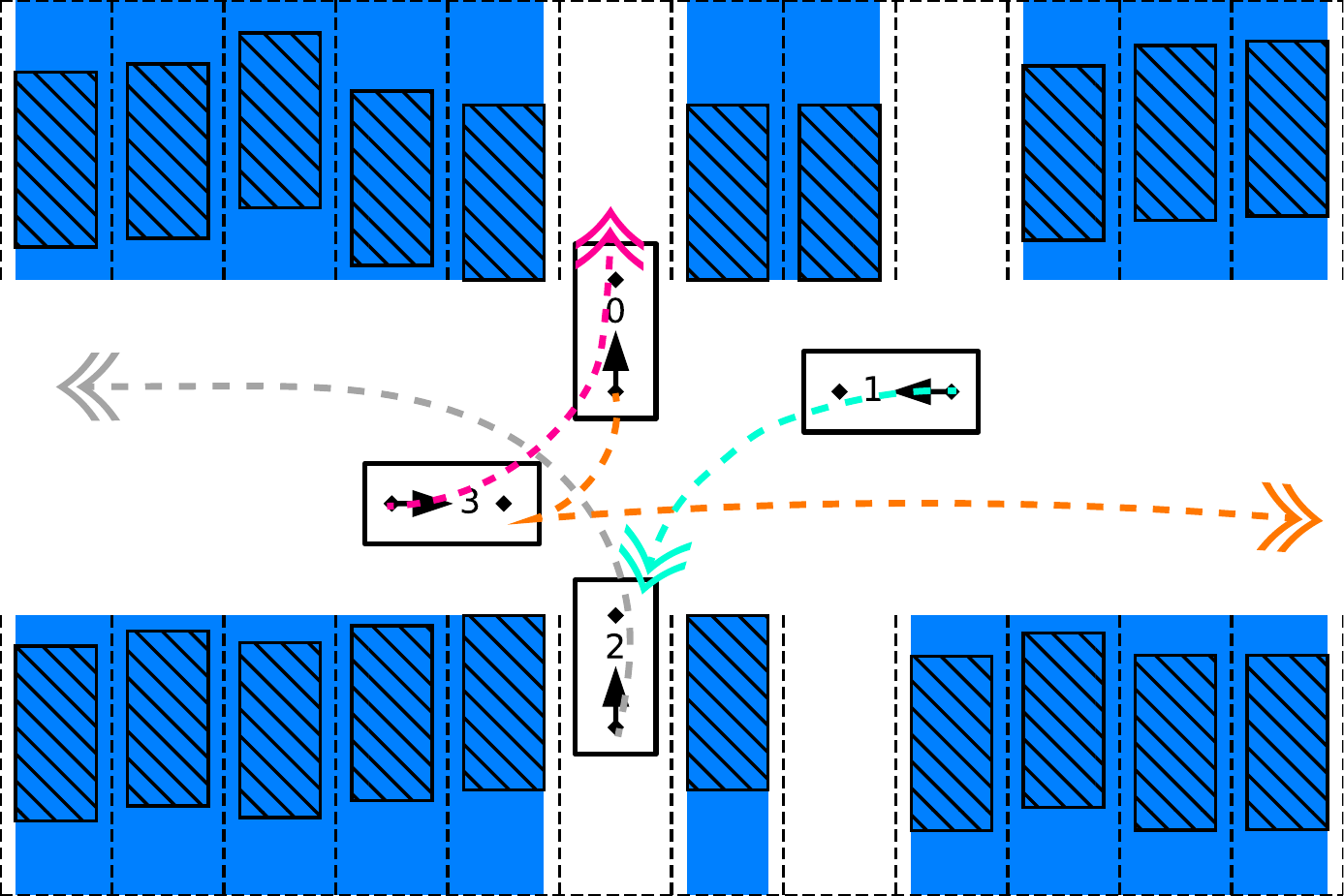}    
\caption{Problem Scenario. The blue regions represent static obstacles in the parking lot, which are the over-approximation of the static vehicles in diagonal hatches. The vehicles 0-3 that we control are plotted with black rectangles, while the short arrows point from the center of the rear axle to the center of the front axle. The dashed curves in orange, cyan, grey, and magenta reflect the intentions of the vehicles.}
\label{fig:scenario}
\end{center}
\end{figure}

The proposed technique is general and not limited to a particular environment, a certain number of vehicles, or vehicle configurations.
For the sake of simplicity, this paper focuses on a specific example: a tightly-constrained parking lot 
represented in Fig.~\ref{fig:scenario}, where four vehicles are involved in a conflict, indexed by $i \in \mathcal{I} = \{0,1,2,3\}$.
Conflicting motion intentions and the lack of free space make it particularly challenging to drive safely and efficiently. 
Vehicles $0$ and $2$ are going to leave their spots for different directions, while vehicles $1$ and $3$ are going to take over the same sets of spots. To resolve this conflict efficiently, the vehicles need to not only avoid collision in close proximity, but also make compromises and determine the passing order so that there is space for everyone to proceed to their destination.
Other scenarios can be found at: bit.ly/rl-cr.

\subsection{Vehicle Modeling}
\label{sec:vehicle-model}
In this scenario, the vehicles operate at low speeds, so tire slip and inertial can be ignored. We therefore model the vehicle dynamics using the kinematic bicycle model
\begin{equation}
    \dot{z} = f(z, u) := \left[
        \begin{matrix}
            \dot{x} \\ \dot{y} \\ \dot{\psi} \\ \dot{v} \\ \dot{\delta_f}
        \end{matrix}
    \right] = \left[
        \begin{matrix}
            v\cos(\psi) \\ v\sin(\psi) \\ \frac{v}{l_{\mathrm{wb}}}\tan(\delta_f) \\ a \\ \omega
        \end{matrix}
    \right], \ u = \left[
        \begin{matrix}
            a \\ \omega
        \end{matrix}
    \right],
    \label{eq:kin_model_ct}
\end{equation}
where the state $z \in \mathcal{Z}$ consists of the position of the center of rear axle $(x,y)$, the heading angle $\psi$, the speed $v$, and the front steering angle $\delta_f$, while the input $u \in \mathcal{U}$ consists of the acceleration $a$ and the steering rate $\omega$. Parameter $l_\mathrm{wb}$ describes the wheelbase.

The geometry of vehicle $i$ is modeled as a polygon
\begin{equation}
    \mathbb{B}(z^{[i]}) := \{ p\in \mathbb{R}^2: G(z^{[i]}) p \leq g(z^{[i]})\} \nonumber,
\end{equation}
where $G(\cdot) \in \mathbb{R}^{4\times2}, g(\cdot) \in \mathbb{R}^4$ can be computed by the vehicle size, position $(x^{[i]},y^{[i]})$ and heading angle $\psi^{[i]}$. Wihtout loss of generality, we assume here that all vehicles share the same body dimension. Similar polyhedral representations also apply to all $M$ static obstacles in this environment
\begin{equation}
    \mathbb{O}^{(m)} := \{ p\in\mathbb{R}^2: A^{(m)} p \leq b^{(m)}\}, \ m = 1, \dots, M, \nonumber.
\end{equation}

\subsection{Optimal Control for Conflict Resolution}

The goal of vehicle $i$ is to reach a state $z^{[i]}_{\mathrm{F}}$ in the terminal set $\mathcal{Z}_{\mathrm{F}}^{[i]}$ 
at the end of its planning horizon $T^{[i]}$ from an initial state $z_0^{[i]}$.

To ensure safety at all time $t$, we require that each vehicle remains collision-free with the static obstacles and all other vehicles. This is expressed as the constraints
\begin{subequations}
    \begin{align}
        \mathrm{dist}\left(\mathbb{B}(z^{[i]}(t)), \mathbb{O}^{(m)}\right) & \geq d_{\mathrm{min}}, \label{eq:collision-avoid-static}\\
        \mathrm{dist}\left(\mathbb{B}(z^{[i]}(t)), \mathbb{B}(z^{[j]}(t))\right) & \geq d_{\mathrm{min}},\label{eq:collision-avoid-vehicle}\\
        \forall t \geq 0, \ i,j \in \mathcal{I}, j \neq i,\ & m = 1,\dots, M \nonumber,
    \end{align}
\end{subequations}
where $d_{\mathrm{min}} > 0$ is a minimum safety distance between each pair of polygons.

We can express the conflict resolution problem as an optimal control formulation:
\begin{subequations}
\label{eq:naive-formulation}
\begin{align}
    \min_{\mathbf{z}^{[i]}, \mathbf{u}^{[i]}, z^{[i]}_{\mathrm{F}}, T^{[i]}} \ & \ \sum_{i \in \mathcal{I}} J^{[i]} = \sum_{i \in \mathcal{I}} \int_{t=0}^{T^{[i]}} c\left(z^{[i]}(t), u^{[i]}(t)\right) dt\nonumber \\
    \text{s.t. } \ & \dot{z}^{[i]}(t) = f(z^{[i]}(t), u^{[i]}(t)), \\
    & z^{[i]}(t) \in \mathcal{Z}, u^{[i]}(t) \in \mathcal{U},\\
    & z^{[i]}(0) = z^{[i]}_0 \\
    & \text{Collision Avoidance Constraints } \eqref{eq:collision-avoid-static}, \eqref{eq:collision-avoid-vehicle} \nonumber,\\
    & z^{[i]}(t) = z^{[i]}_{\mathrm{F}} \in \mathcal{Z}_{\mathrm{F}}^{[i]}, \forall t \geq T^{[i]}, \\
    & \forall i \in \mathcal{I}, t \geq 0, \nonumber
\end{align}
\end{subequations}
where $c(\cdot, \cdot)$ is the stage cost. However, directly solving problem \eqref{eq:naive-formulation} is intractable due to the following reasons:
\begin{enumerate}[label=(\roman*)]
    \item Discretization in time is challenging since the horizon $T^{[i]}$ is an optimization variable;
    \item Discretization in space around a trajectory is also hard as there is no feasible initial reference available;
    \item Constraints \eqref{eq:collision-avoid-static} and \eqref{eq:collision-avoid-vehicle} are non-differentiable and therefore not amendable for use with existing gradient- and Hessian-based solvers.
\end{enumerate}

In Section~\ref{sec:rl} and Section~\ref{sec:planning} to follow, we will elaborate on the details of our approach to addressing these issues.

\section{RL-based Conflict Resolution in Grids}
\label{sec:rl}

Since conflict resolution in tight spaces requires complex maneuvers, obtaining sequences of ``tactical'' configurations for vehicles is crucial so that the optimization problem~\eqref{eq:naive-formulation} can be guided towards feasible solutions. In this work, these sequences are called ``strategies'' and are solved by multi-agent reinforcement learning (RL) in a simplified environment.

\subsection{Partially Observable Markov Decision Process}
\label{sec:ma-pomdp}

A multi-agent partially observable Markov decision process (MA-POMDP) is defined by a state space $\mathcal{S}$ for the possible configurations of all vehicles, action space $\mathcal{A}$, and observation space $\mathcal{O}$ for each vehicle $i \in \mathcal{I}$. All vehicles use a single shared policy $\pi_{\theta}: \mathcal{O} \mapsto \mathcal{A}$ to choose action, which produces the next state of all vehicles according to the state transition function $\mathcal{T}: \mathcal{S} \times \Pi_{i \in \mathcal{I}} \mathcal{A} \mapsto \mathcal{S}$. Each vehicle $i$ obtains its private observation $o^{[i]}$ with the observation function $\Omega^{[i]}: \mathcal{S} \mapsto \mathcal{O}$ and receives rewards $r^{[i]}$ by $R^{[i]}: \mathcal{S} \times \mathcal{A} \mapsto \mathbb{R}$. Each vehicle $i$ aims to maximize its own expected reward $R^{[i]} = \sum_{k=0}^{K^{[i]}} \gamma^{k} (r^{[i]})^{k}$, where $\gamma$ is a discount factor and $K^{[i]}$ is the number of steps that vehicle $i$ takes to reach its destination.

\subsubsection{State Space}
\begin{figure}
	\centering
	\begin{subfigure}[t]{0.59\columnwidth}
		\centering
		\includegraphics[width=\textwidth]{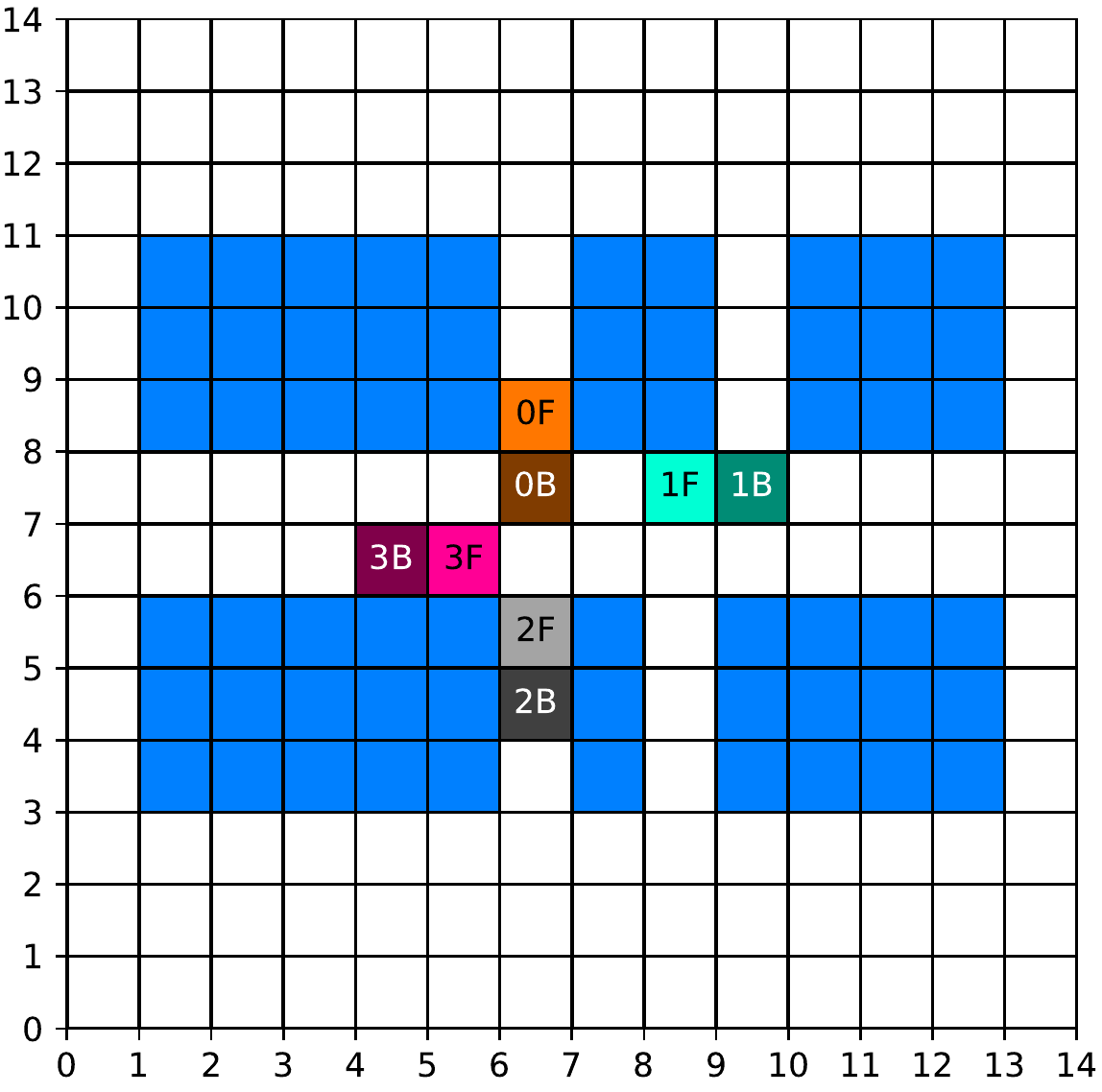}
	    \caption{Grid Map}
	    \label{fig:grid-map}
	\end{subfigure}%
	~
	\begin{subfigure}[t]{0.39\columnwidth}
		\centering
		\includegraphics[width=\textwidth]{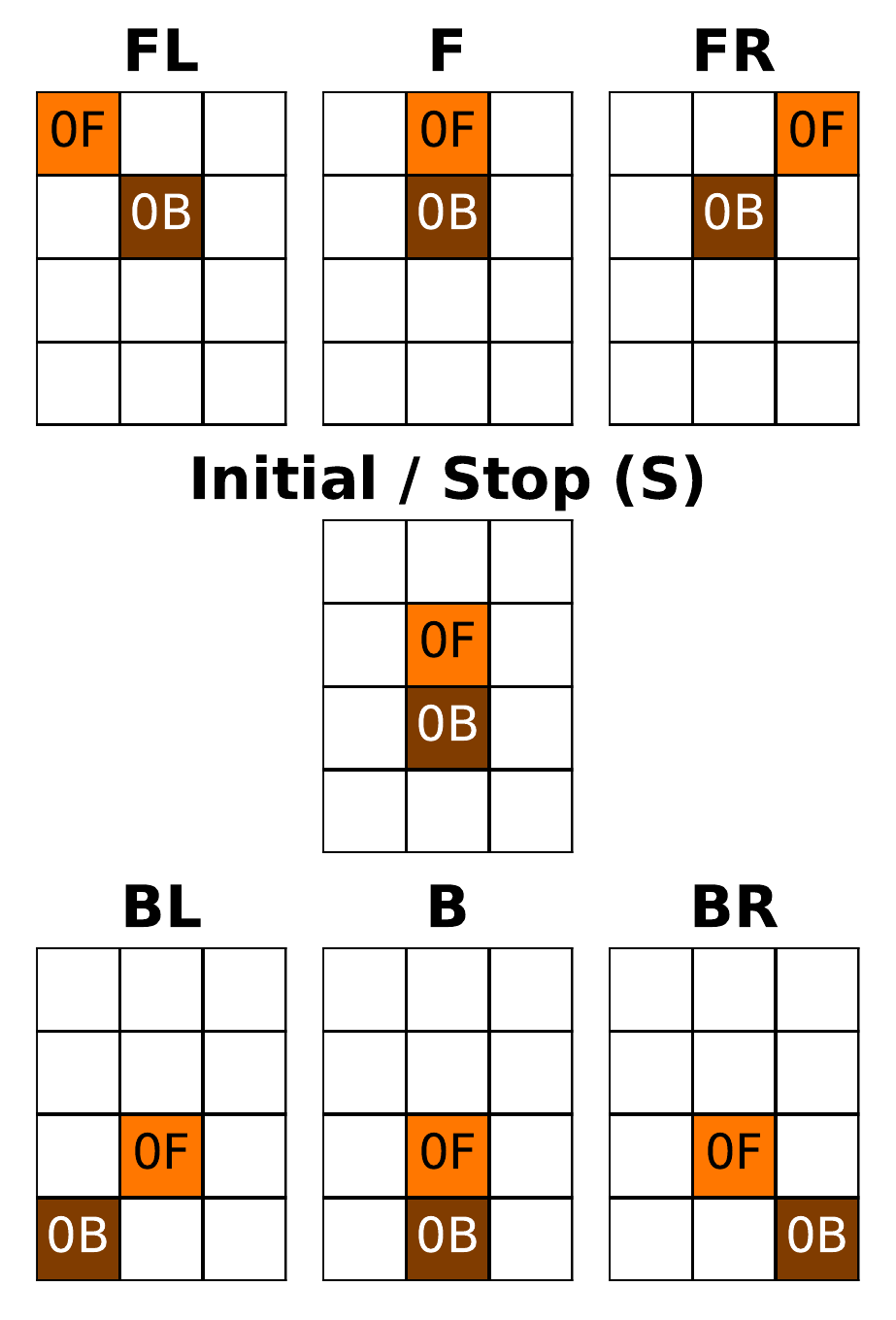}
		\caption{Dynamics in grids}
		\label{fig:grid-dynamics}
	\end{subfigure}
	\caption{Vehicles in the discretized environment. (a) The conflict scenario. The blue grids are occupied by static obstacles. The ``0F'' grid [6,8] in orange is the front of vehicle 0, and the ``0B'' grid [6,7] in dark orange is the back of vehicle 0. Similar annotation applies to vehicle 1,2,3 in cyan, grey, and magenta respectively. (b) Single-vehicle dynamics in a free grid map.}
\end{figure}

We take the environment in Fig.~\ref{fig:scenario} and discretize it into a grid map as shown in Fig.~\ref{fig:grid-map}. The grid state $s^{[i]} \in \mathbb{Z}^{2\times2}_{+}$ of each vehicle $i$ is described by the grid coordinates of its ``front'' and ``back'' cells. The rationale of this design is to keep the dynamics and geometry of the vehicle in the grid map similar to those in the continuous environment described in Section~\ref{sec:vehicle-model}.

\subsubsection{Action Space and State Transition}
The action space $\mathcal{A}$ contains seven discrete actions: \{Stop (S), Forward (F), Forward Left (FL), Forward Right (FR), Backward (B), Backward Left (BL), Backward Right (BR)\}, and the single-vehicle dynamics in a free grid map is demonstrated in Fig.~\ref{fig:grid-dynamics}. It can be observed that this grid-version vehicle dynamics also follows the nonholonomic constraints~\eqref{eq:kin_model_ct} so that it is impossible to turn on the spot and move sideways. 

In the multi-agent environment, all vehicles will take action and move simultaneously. If there are collisions among vehicles or against static obstacles, the corresponding vehicles will be ``bounced back'' so that their grid states remain unchanged at the next step. However, the collision will still be recorded to calculate reward.

\subsubsection{Reward Function}
The policy $\pi$ aims to drive each vehicle to its destination as quickly as possible while maintaining collision-free. Therefore, the reward function of vehicle $i$ is
\begin{equation}
    \begin{split}
        R^{[i]}(s^{[i]},a^{[i]}) = & r_{\mathrm{c}} \mathbbm{1}(\mathrm{collision}) + r_{\mathrm{s}}\mathbbm{1}(a^{[i]} = \mathrm{Stop}) \\
        + & r_{\mathrm{d}} d(s^{[i]}, \eta^{[i]}) + r_{\mathrm{t}} + r_{\eta}\mathbbm{1}(s^{[i]} = \eta^{[i]}),
    \end{split}
\end{equation}
where $r_{\mathrm{c}} <0$ penalizes the collision with any other vehicle or static obstacle, $r_{\mathrm{s}} <0$ penalizes the action of stopping, $r_{\mathrm{d}}<0$ penalizes the distance away from its destination $g^{[i]}$, $r_{\mathrm{t}}<0$ penalizes time consumption, and $r_{\eta} >0$ provides incentive for reaching the destination. The destination $\eta^{[i]}\in \mathbb{Z}^{2\times2}_{+}$ is represented by the target cells for the ``front'' and ``back'' of the vehicle and the distance is computed by $d(s,\eta) = \| s-\eta\|_{\mathrm{F}}$, where $\| \cdot\|_{\mathrm{F}}$ is the Frobenius norm.

\subsubsection{Observation}
\label{sec:rl-observation}
\begin{figure}
	\centering
	\begin{subfigure}[t]{0.481\columnwidth}
		\centering
		\includegraphics[width=\textwidth]{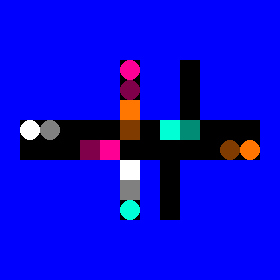}
	    \caption{Observation of vehicle 0}
	    \label{fig:obs-vehicle0-4v}
	\end{subfigure}%
	~
	\begin{subfigure}[t]{0.481\columnwidth}
		\centering
		\includegraphics[width=\textwidth]{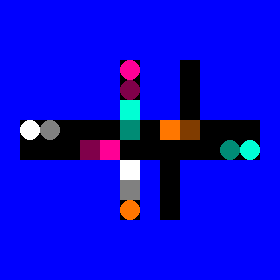}
		\caption{Observation of vehicle 1}
		\label{fig:obs-vehicle1-4v}
	\end{subfigure}
	\caption{Private observations of vehicle $i = 0,1$ at the same step $k=0$. Black regions are the free spaces, and blue squares describe static obstacles. Orange squares represent vehicle $i$ itself, and cyan, white, and magenta squares represent other vehicles. The circles in the corresponding color are their destinations, which could be occluded by vehicle bodies.}
    \label{fig:rl-observation}
\end{figure}

The observation space $\mathcal{O}$ is $I \times I$ RGB images of the entire parking plot as in Fig.~\ref{fig:rl-observation}. The observation functions $\Omega^{[i]}$ are designed so that among all colors used for describing vehicles, the body and destination of vehicle $i$ itself always use normal/dark orange in its private observation $o^{[i]}$.

\subsection{Multi-agent Reinforcement Learning}
Since all vehicles $i \in \mathcal{I}$ have identical action spaces, observation spaces,
and reward structures, the vehicles are homogeneous, so their policy can be trained efficiently with parameter sharing,~\cite{gupta_cooperative_2017, terry_parameter_2022}.

We use deep Q-network (DQN) with convolutional layers to approximate the observation-action value function $Q(o, a;\theta)$, and the shared policy is $\pi_{\theta} = \operatorname{argmax}_{a} Q(o, a \mid \theta)$. The experiences of all vehicles are stored in the replay buffer $\mathcal{D}$ to train the policy simultaneously. An experience sample $e = \left(o, a, r, o^{\prime}\right) \in \mathcal{D}$ could come from any vehicle $i \in \mathcal{I}$ and consists of its current observation $o$, action $a$,  reward $r$, and observation at the next step $o^{\prime}$. The optimal parameters $\theta^{*}$ of the DQN policy are obtained by minimizing the loss $\ell$ based on the temporal difference:
\begin{equation}
    \small
    \ell(\theta) = \mathbb{E}_{(o,a,r,o^{\prime}) \sim \mathcal{D}} \left[ \left( r + \gamma \max_{a^{\prime}} Q(o^{\prime}, a^{\prime}; \theta_{\_}) - Q(o, a;\theta)\right)^2 \right],
\end{equation}
where $\theta_{\_}$ defines a fixed target network to be updated periodically with new parameters $\theta$.

Although the training is centralized with the collected experiences of all vehicles, the execution is decentralized since the trained policy $\pi_{\theta^{*}}$ is a function of private observation only. Moreover, the shared policy still permits diverse behavior among vehicles because, for the same scenario, the observation functions $\Omega^{[i]}$ generate unique observations for each vehicle $i$, as illustrated in Fig.~\ref{fig:rl-observation}.

\begin{figure}
\begin{center}
\includegraphics[width=\linewidth]{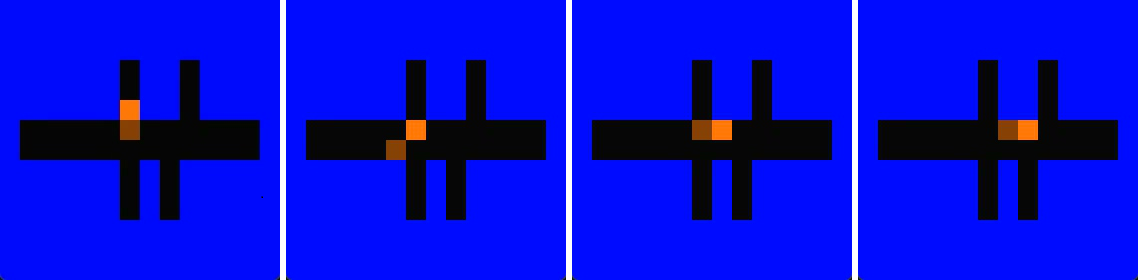}    
\caption{Grid states of a vehicle at four consecutive steps.}
\label{fig:rl-result-1v}
\end{center}
\end{figure}

By running the policy $\pi_{\theta^{*}}$, we can get the strategies $\mathbf{s}^{[i]} = \{s^{[i]}_k \mid k=0,\dots, K^{[i]}\}$ for vehicles $i \in \mathcal{I}$, which record the steps they take to resolve the conflict in the discrete environment. An example is illustrated in Fig.~\ref{fig:rl-result-1v}.

\section{Multi-vehicle Trajectory Planning}
\label{sec:planning}
In this section, we will introduce our approach to leverage the strategies $\mathbf{s}^{[i]}, i \in \mathcal{I}$ provided by RL-based conflict resolution in the discrete environment to solve problem~\eqref{eq:naive-formulation}.

\subsection{From Grids to Constraints}
For each vehicle $i$, the strategy $\mathbf{s}^{[i]}$ describes a sequence of tactical vehicle configurations in the simplified grid environment at discrete time steps. These configurations are then transformed as a set of constraints for the optimal control problem.

Given the grid resolution $L$ and the grid state of a vehicle $s^{[i]}_k = \left[[X^{[i]}_{\mathrm{F},k}, Y^{[i]}_{\mathrm{F},k}], [X^{[i]}_{\mathrm{B}, k}, Y^{[i]}_{\mathrm{B}, k}]\right]$, the front and back cells in the continuous ground coordinates become square convex sets $\Bar{\mathcal{Z}}_{\mathrm{F}, k}^{[i]}, \Bar{\mathcal{Z}}_{\mathrm{B}, k}^{[i]}$:
\begin{subequations}
\small
\begin{align}
    & \Bar{\mathcal{Z}}_{\mathrm{F}, k}^{[i]} = \mathbf{\mathrm{conv}}\left( V^{[i]}_{\mathrm{F},k} \right), V^{[i]}_{\mathrm{F},k} = L\left[
    \begin{matrix}
            X^{[i]}_{\mathrm{F},k} & Y^{[i]}_{\mathrm{F},k} \\
            X^{[i]}_{\mathrm{F},k} & Y^{[i]}_{\mathrm{F},k}+1 \\
            X^{[i]}_{\mathrm{F},k}+1 & Y^{[i]}_{\mathrm{F},k}+1 \\
            X^{[i]}_{\mathrm{F},k}+1 & Y^{[i]}_{\mathrm{F},k}
    \end{matrix}
    \right], \\
    & \Bar{\mathcal{Z}}_{\mathrm{B}, k}^{[i]} = \mathbf{\mathrm{conv}}\left( V^{[i]}_{\mathrm{B},k} \right), V^{[i]}_{\mathrm{B},k} =  L\left[
    \begin{matrix}
            X^{[i]}_{\mathrm{B},k} & Y^{[i]}_{\mathrm{B},k} \\
            X^{[i]}_{\mathrm{B},k} & Y^{[i]}_{\mathrm{B},k}+1 \\
            X^{[i]}_{\mathrm{B},k}+1 & Y^{[i]}_{\mathrm{B},k}+1 \\
            X^{[i]}_{\mathrm{B},k}+1 & Y^{[i]}_{\mathrm{B},k}
    \end{matrix}
    \right],
\end{align}
\end{subequations}
where $\mathbf{\mathrm{conv}}(\cdot)$ denotes the convex hull of vertices.

\begin{figure}
\begin{center}
\includegraphics[width=0.97\linewidth]{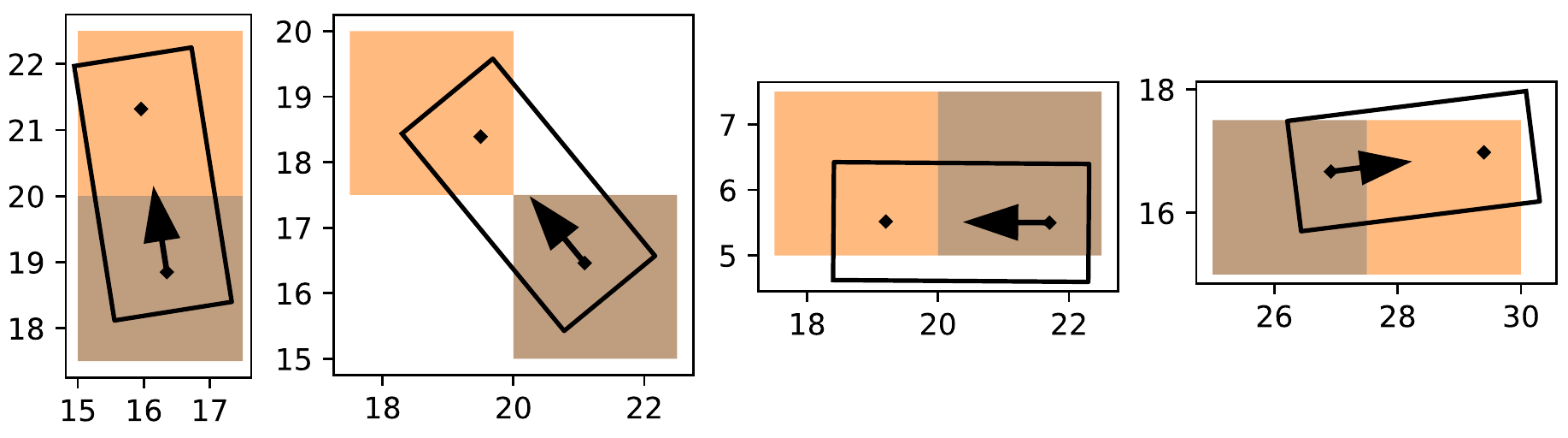}    
\caption{Feasible vehicle configurations under strategy-guided constraints. Each dot with an arrow indicates the center of the rear axle, while each dot without an arrow is the center of the front axle. The normal and dark orange squares represent the constraint sets $\Bar{\mathcal{Z}}_{\mathrm{F}}, \Bar{\mathcal{Z}}_{\mathrm{B}}$ respectively.}
\label{fig:grid-constriants}
\end{center}
\end{figure}

We denote by $T_{\mathrm{s}}$ the sampling time between two discrete steps $k$ and $k+1$, and enforce configuration constraints so that at time $t = kT_{\mathrm{s}}$, the center of the back and the front axle should be inside the convex sets $\Bar{\mathcal{Z}}_{\mathrm{F}, k}^{[i]}, \Bar{\mathcal{Z}}_{\mathrm{B}, k}^{[i]}$, formally:
\begin{subequations}
    \label{eq:strategy-guided-config-constraints}
    \begin{align}
        \left[ x^{[i]}(kT_{\mathrm{s}}), y^{[i]}(kT_{\mathrm{s}})\right]^{\top} & \in \Bar{\mathcal{Z}}_{\mathrm{B}, k}^{[i]} \\
        \left[ x_{\mathrm{F}}^{[i]}(kT_{\mathrm{s}}), y_{\mathrm{F}}^{[i]}(kT_{\mathrm{s}})\right]^{\top} & \in \Bar{\mathcal{Z}}_{\mathrm{F}, k}^{[i]}, \\
        \forall k = 0, \dots, K^{[i]}, i &\in \mathcal{I}, \nonumber
    \end{align}
    where $K^{[i]}$ is the number of steps that
vehicle $i$ takes to reach its destination, and the center of front axle $(x_{\mathrm{F}}^{[i]}, y_{\mathrm{F}}^{[i]})$ is computed by
    \begin{equation}
    x_{\mathrm{F}}^{[i]} = x^{[i]} + l_{\mathrm{wb}} \cos(\psi^{[i]}), y_{\mathrm{F}}^{[i]} = y^{[i]} + l_{\mathrm{wb}} \sin(\psi^{[i]}).
    \end{equation}
\end{subequations}
We call~\eqref{eq:strategy-guided-config-constraints} strategy-guided constraints and Fig.~\ref{fig:grid-constriants} shows some examples of feasible vehicle configurations.

\subsection{Strategy-guided Optimal Control}
The strategy-guided constraints~\eqref{eq:strategy-guided-config-constraints} indicate the combinatorial thinking for vehicles to resolve the conflict, i.e. the vehicles need to reach certain tactical configurations at ``key'' time steps. 
Moreover, the planning horizon can be rewritten as $T^{[i]} = T_{\mathrm{s}}K^{[i]}$ so that the time discretization can be easily applied based on the number of steps $K^{[i]}$, and $T_{\mathrm{s}}$ is a single optimization variable for all vehicles.

The non-differentiable collision avoidance constraints~\eqref{eq:collision-avoid-static} and \eqref{eq:collision-avoid-vehicle} can be reformualted as smooth nonlinear constraints with the method provided by~\cite{zhang_optimization-based_2020}:
\begin{subequations}
    \label{eq:obca-constr-static}
    \begin{align}
        \mathrm{dist}\left(\mathbb{B}(z^{[i]}), \mathbb{O}^{(m)}\right) \geq d_{\mathrm{min}} & \Leftrightarrow \nonumber\\
        \exists \lambda^{[i, m]} \geq 0, \mu ^{[i, m]} \geq 0, \|A^{(m)\top}\lambda^{[i, m]}\| & = 1 \\
        g^{\top}(0) \mu^{[i, m]}+\left(A^{(m)} p(z^{[i]})-b^{(m)}\right)^{\top} \lambda^{[i, m]} & \geq d_{\mathrm{min}} \\
        G^{\top}(0) \mu^{[i, m]} + R(z^{[i]})^{\top} A^{(m) \top} \lambda^{[i, m]} & = 0
    \end{align}
\end{subequations}
\begin{subequations}
    \label{eq:obca-constr-vehicle}
    \begin{align}
        \mathrm{dist}\left(\mathbb{B}(z^{[i]}), \mathbb{B}(z^{[j]})\right) \geq d_{\mathrm{min}} & \Leftrightarrow \nonumber\\
        \exists \lambda^{[i, j]} \geq 0, \mu ^{[i, j]} \geq 0, \|G^{[j]}(z^{[j]})^{\top}\lambda^{[i, j]}\| & = 1 \\
        g^{\top}(0) \mu^{[i, j]}+\left(G(z^{[j]}) p(z^{[i]})-g(z^{[j]})\right)^{\top} \lambda^{[i, j]} & \geq d_{\mathrm{min}} \\
        G^{\top}(0) \mu^{[i, j]} + R(z^{[i]})^{\top} A^{(m) \top} \lambda^{[i, m]} & = 0 \\
        \forall \ i,j \in \mathcal{I}, j \neq i,\ m = 1,\dots, M & \nonumber,
    \end{align}
\end{subequations}
where $p(z^{[i]}) = [x^{[i]}, y^{[i]}]^{\top}$ is the position of the vehicle, and $R(z^{[i]})$ is the rotation matrix that depends on the heading angle $\psi^{[i]}$. For conciseness, we omit time $t$ above as the function argument of states.

The optimal control problem~\eqref{eq:naive-formulation} can now be reformulated and augmented as:
\begin{subequations}
\label{eq:complete-formulation}
\begin{align}
    \min_{\substack{\mathbf{z}^{[i]}, \mathbf{u}^{[i]}, z^{[i]}_{\mathrm{F}} \\ \lambda^{[i, \cdot]}, \mu^{[i, \cdot]}, T_{\mathrm{s}}}} \ & \ \sum_{i \in \mathcal{I}} J^{[i]} = \sum_{i \in \mathcal{I}} \int_{t=0}^{T_{\mathrm{s}}K^{[i]}} c\left(z^{[i]}(t), u^{[i]}(t)\right) dt\nonumber \\
    \text{s.t. } \ & \dot{z}^{[i]}(t) = f(z^{[i]}(t), u^{[i]}(t)), \\
    & z^{[i]}(t) \in \mathcal{Z}, u^{[i]}(t) \in \mathcal{U},\\
    & z^{[i]}(0) = z^{[i]}_0, \\
    & \text{Strategy-guided Configuration Constraints } \eqref{eq:strategy-guided-config-constraints} \nonumber,\\
    & \text{Collision Avoidance Constraints } \eqref{eq:obca-constr-static}, \eqref{eq:obca-constr-vehicle} \nonumber,\\
    & z^{[i]}(t) = z^{[i]}_{\mathrm{F}} \in \mathcal{Z}_{\mathrm{F}}^{[i]}, \forall t \geq T_{\mathrm{s}}K^{[i]}, \\
    & \forall i\in \mathcal{I}, t \geq 0. \nonumber
\end{align}
\end{subequations}

\subsection{Progressive Steps for Initial Guess}
\label{sec:hierarchical-steps}
Although problem~\eqref{eq:complete-formulation} is differentiable, can be discretized in time, and is guided by the learned configuration constraints, it is still a non-convex and nonlinear programming (NLP) problem, which is numerically challenging to solve in general. The solvers require good initial guesses to find local optima. Here we provide a hierarchy to obtain near-optimal initial guesses progressively.

\subsubsection{Single-vehicle Pose Interpolation}
For each vehicle $i$, we can get a set of nominal poses $\left\{ (\Bar{x}^{[i]}_k, \Bar{y}^{[i]}_k, \Bar{\psi}^{[i]}_k)\right\}$ by assuming the vehicle is at the centers of the strategy-guided configuration sets $\left\{ \Bar{\mathcal{Z}}_{\mathrm{F}, k}^{[i]}, \Bar{\mathcal{Z}}_{\mathrm{B}, k}^{[i]}\right\}$. A Bézier curve interpolation is analytically computed between every two nominal poses with control points to force tangents.

\subsubsection{Single-vehicle Strategy-guided Trajectory}
For each vehicle $i$, a single-vehicle optimal control problem is set up as:
\begin{subequations}
\label{eq:single-vehicle-staet-intput-ws}
\begin{align}
    \min_{\mathbf{z}^{[i]}, \mathbf{u}^{[i]}, z^{[i]}_{\mathrm{F}}} \ & \ J^{[i]} = \int_{t=0}^{\Bar{T}_{\mathrm{s}}K^{[i]}} c\left(z^{[i]}(t), u^{[i]}(t)\right) dt\nonumber \\
    \text{s.t. } \ & \dot{z}^{[i]}(t) = f(z^{[i]}(t), u^{[i]}(t)), \\
    & z^{[i]}(t) \in \mathcal{Z}, u^{[i]}(t) \in \mathcal{U},\\
    & z^{[i]}(0) = z^{[i]}_0 \\
    & \text{Strategy-guided Configuration Constraints } \eqref{eq:strategy-guided-config-constraints} \nonumber,\\
    & z^{[i]}(\Bar{T}_{\mathrm{s}}K^{[i]}) = z^{[i]}_{\mathrm{F}}\in \mathcal{Z}_{\mathrm{F}}^{[i]}, \\
    & \forall 0 \leq t \leq \Bar{T}_{\mathrm{s}}K^{[i]}, \nonumber
\end{align}
\end{subequations}
where a constant parameter $\Bar{T}_{\mathrm{s}}$ is given as the sampling time. By solving problem~\eqref{eq:single-vehicle-staet-intput-ws}, we can obtain a strategy-guided trajectory $\Tilde{\mathbf{z}}^{[i]}$ and input profile $\Tilde{\mathbf{u}}^{[i]}$ that are kinematically feasible and compliant with the strategy-guided configurations. Note that the collision avoidance constraints~\eqref{eq:obca-constr-static},~\eqref{eq:obca-constr-vehicle} are not considered at this step.

\subsubsection{Single-vehicle Collision-free Trajectory}
With the strategy-guided trajectory $\Tilde{\mathbf{z}}^{[i]}$ and input $\Tilde{\mathbf{u}}^{[i]}$, we can firstly warm start the auxiliary variables $\lambda^{[i, m]}, \mu^{[i, m]}$ introduced in~\eqref{eq:obca-constr-static}, using the approach provided by~\cite{zhang_autonomous_2019}. Then, we add constraints~\eqref{eq:obca-constr-static} to problem~\eqref{eq:single-vehicle-staet-intput-ws} to obtain single-vehicle collision-free trajectory $\Bar{\mathbf{z}}^{[i]}, \Bar{\mathbf{u}}^{[i]}$ that is both compliant with strategy-guided configurations and collision-free against static obstacles. Optimal auxiliary variables are also recorded as $\{\Bar{\lambda}^{[i, m]}, \Bar{\mu}^{[i, m]}\}$.

\subsubsection{Multi-vehicle Conflict Resolution}
Similar as the step above, the dual variables $\lambda^{[i, j]}, \mu^{[i, j]}$ in the inter-vehicle constraints~\eqref{eq:obca-constr-vehicle} can be warm started as $\Bar{\lambda}^{[i, j]}, \Bar{\mu}^{[i, j]}$ by $\Bar{\mathbf{z}}^{[i]}$ from all vehicles. 
Finally, we solve problem~\eqref{eq:complete-formulation} with the initial guesses $\Bar{\mathbf{z}}^{[i]}, \Bar{\mathbf{u}}^{[i]}, \Bar{\mathbf{\lambda}}^{[i, m]}, \Bar{\mathbf{\mu}}^{[i, m]}, \Bar{\lambda}^{[i, j]}, \Bar{\mu}^{[i, j]}, \Bar{T}_\mathrm{s}, \forall i, j \in \mathcal{I}, i \neq j, m = 1, \dots, M$.

\section{Results}
\label{sec:results}

In this section, we present the numerical details of the conflict resolution scenario and the simulation results.  The source code and demo video can be found at: 
bit.ly/rl-cr.

The parking lot region presented in Fig.~\ref{fig:scenario} is of size $30 \mathrm{m} \times 20 \mathrm{m}$. Other states and inputs of vehicles $i \in \mathcal{I}$ are constrained by $v^{[i]} \in [-2.5, 2.5] \mathrm{m/s}$, $\delta_{f}^{[i]} \in [-0.85, 0.85] \mathrm{rad}$, $a^{[i]} \in [-1.5, 1.5] \mathrm{m/s^2}$, $w^{[i]} \in [-1, 1] \mathrm{rad/s}$. 
The vehicle body polygons $\mathbb{B}(z^{[i]})$ are rectangles with length $3.9 \mathrm{m}$ and width $1.8 \mathrm{m}$. The wheelbase of the vehicles is $2.5 \mathrm{m}$.

 The initial poses of $z_0^{[i]}$ and the ranges of final poses in $\mathcal{Z}_{\mathrm{F}}^{[i]}$ are reported in Table.~\ref{tab:initial-final-pose}. All other state and input components are $0$ at the initial and final time steps.

\begin{table}[t]
\centering
\caption{Constraints for Initial and Final Poses}
\label{tab:initial-final-pose}
\begin{tabular}{ccccc}
\toprule
$i$ & Initial Pose $(x,y,\psi)$ & Final $x$ & Final $y$ & Final $\psi$ \\ \midrule
0 & $(16.25, 18.75, \frac{1}{2}\pi)$  & $[27.5, 30]$ & $[15, 17.5]$ & $[-\frac{1}{10}\pi, \frac{1}{10}\pi]$ \\
\midrule
1 & $(23.75, 18.75, \pi)$  & $[15, 17.5]$ & $[10, 12.5]$ & $[-\frac{3}{5}\pi, -\frac{2}{5}\pi]$ \\       
\midrule
2 & $(16.25, 11.25, \frac{1}{2}\pi)$  & $[5, 7.5]$ & $[17.5, 20]$ & $[\frac{9}{10}\pi, \frac{11}{10}\pi]$ \\   
\midrule
3 & $(11.25, 16.25, 0)$  & $[15, 17.5]$ & $[22.5, 25]$ & $[\frac{2}{5}\pi, \frac{3}{5}\pi]$
\end{tabular}
\end{table}

\subsection{RL-based Conflict Resolution in Grids}

\begin{figure}
\begin{center}
\includegraphics[width=0.97\linewidth]{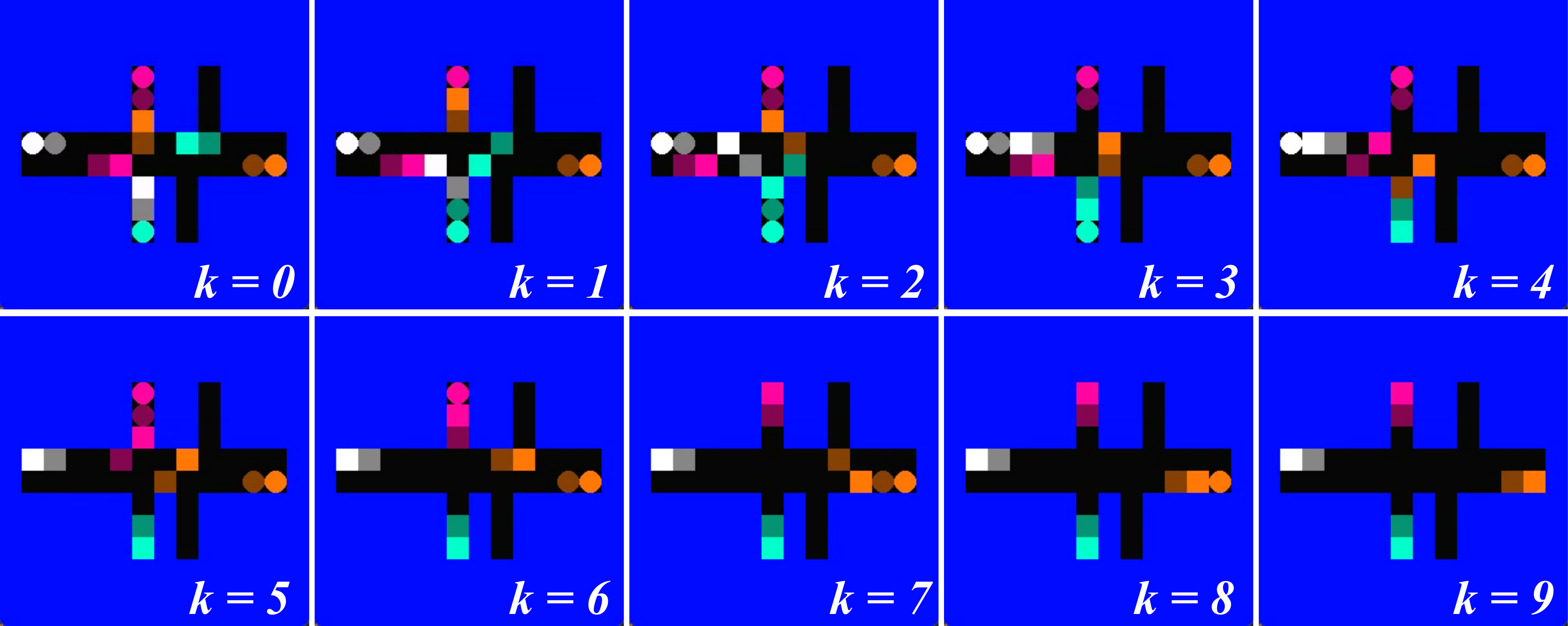}    
\caption{Steps taken by the trained DQN policy to resolve the conflict in grids.}
\label{fig:rl-results}
\end{center}
\end{figure}

\begin{figure}
\begin{center}
\includegraphics[width=0.97\linewidth]{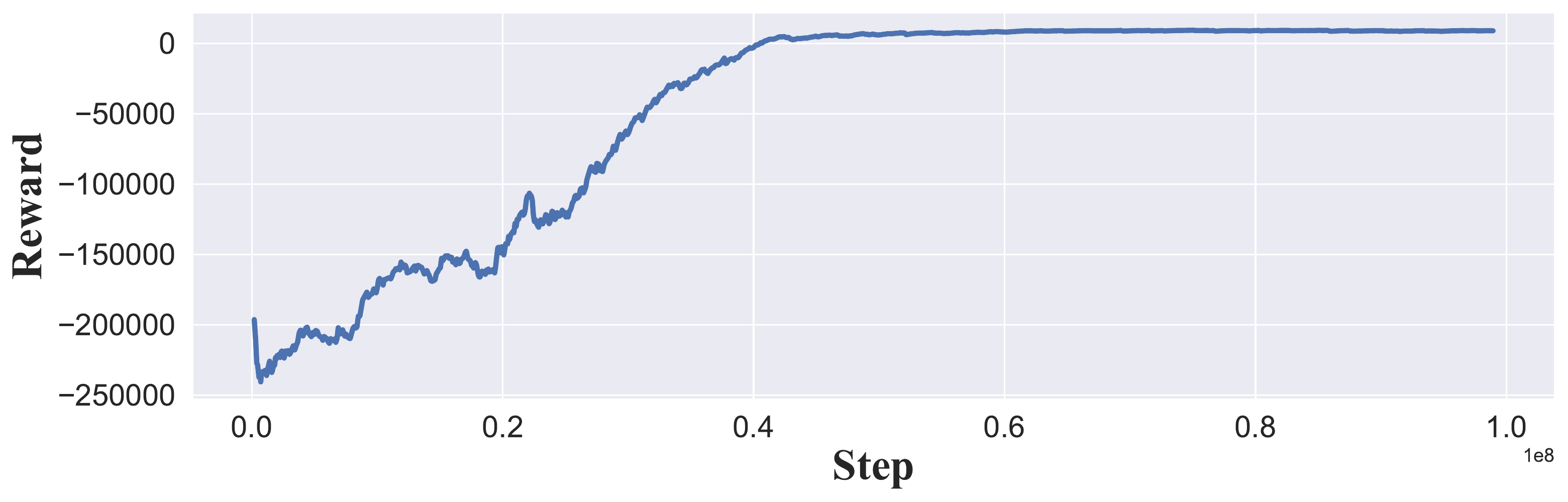}    
\vspace{-0.5em}
\caption{Smoothed episode reward during training.}
\label{fig:training_rewards}
\end{center}
\end{figure}

The grid resolution is $L=2.5 \mathrm{m}$ when creating the discrete grid map and the observation images are of size 140 $\times$ 140. The coefficients of the reward function are: $r_\mathrm{c} = -10^3, r_\mathrm{s}=-10, r_\mathrm{d}=r_\mathrm{t}=-1, r_\mathrm{\eta}=10^4$. The PettingZoo library,~\cite{terry_pettingzoo_2021}, is used for creating the multi-agent RL environment, and the DQN algorithm is implemented with the default CNN policy from Stable-Baselines3,~\cite{ran_stable-baselines3_2021}. The size of the replay buffer $\mathcal{D}$ is $10^5$. In the first $70\%$ of the training period, the learning rate decreases from $5\times10^{-3}$ to $2.5\times10^{-4}$, and the $\epsilon$-greedy probability decreases from $1$ to $0.2$. The algorithm is trained for $10^{8}$ steps, and the episode reward in~Fig.~\ref{fig:training_rewards} is smoothed by the exponentially weighted window method with a smoothing factor of 0.92. The reward at convergence reflects that 
the trained policy can drive the vehicles to their destinations efficiently without collision penalties.

In the runtime, each vehicle takes action based on the trained DQN model $a^{[i]}_k = \pi_{\theta^{*}}(o^{[i]}_k)$ and the resulting strategies $\mathbf{s}^{[i]}$ are shown in Fig.~\ref{fig:rl-results}. The following observations can be made:
\begin{enumerate}[label=(\roman*)]
    \item The model has learned to drive the vehicles to their destinations with the correct heading angle, since the target cells for both the front and the back of vehicles are provided explicitly. During steps $k=2\sim4$, the orange vehicle 0 firstly backs up from the spot on top, then changes its direction with a spot at the bottom.
    \item The model has learned to make compromises for the ``social good''. To create more room in the crowded intersection, the orange vehicle 0 pulls forward into the spot on top at time $k=1$, and the magenta vehicle 3 backs up along the lane at time $k=1,2$. Although these actions will slightly lower their own rewards, they prevent everyone from getting stuck or receiving huge collision penalties.
    \item The model still generates diverse behavior despite being shared with all vehicles. Vehicle 1 and 2 drive directly towards their destinations while vehicle 0 and 3 makes compromises as described above.
\end{enumerate}

\subsection{Startegy-guided Trajectory Planning}

\begin{figure}
	\centering
	\begin{subfigure}[t]{0.49\columnwidth}
		\centering
		\includegraphics[width=\textwidth]{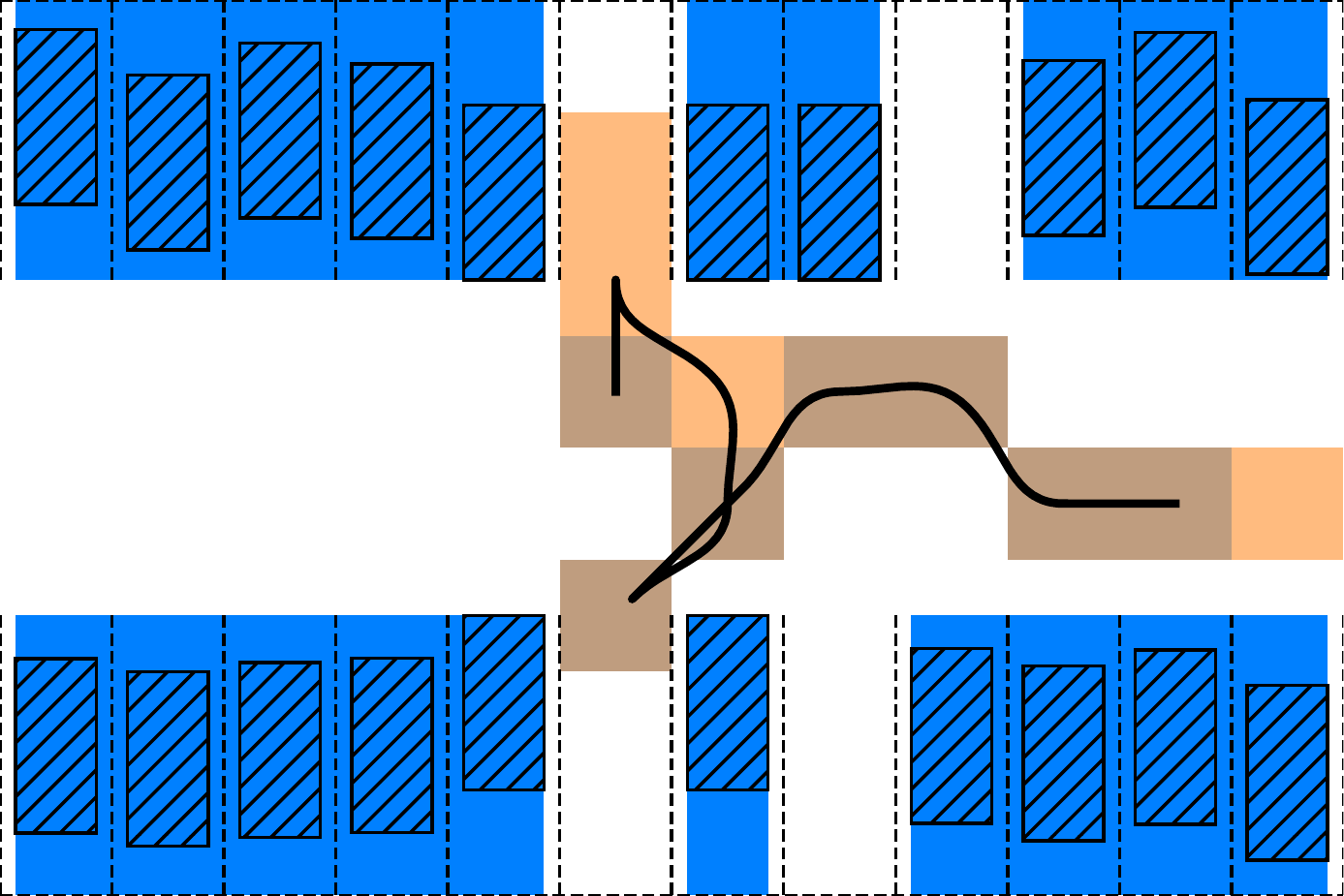}
	    \caption{Pose interpolation}
	    \label{fig:spline-v0}
	\end{subfigure}%
	~
	\begin{subfigure}[t]{0.49\columnwidth}
		\centering
		\includegraphics[width=\textwidth]{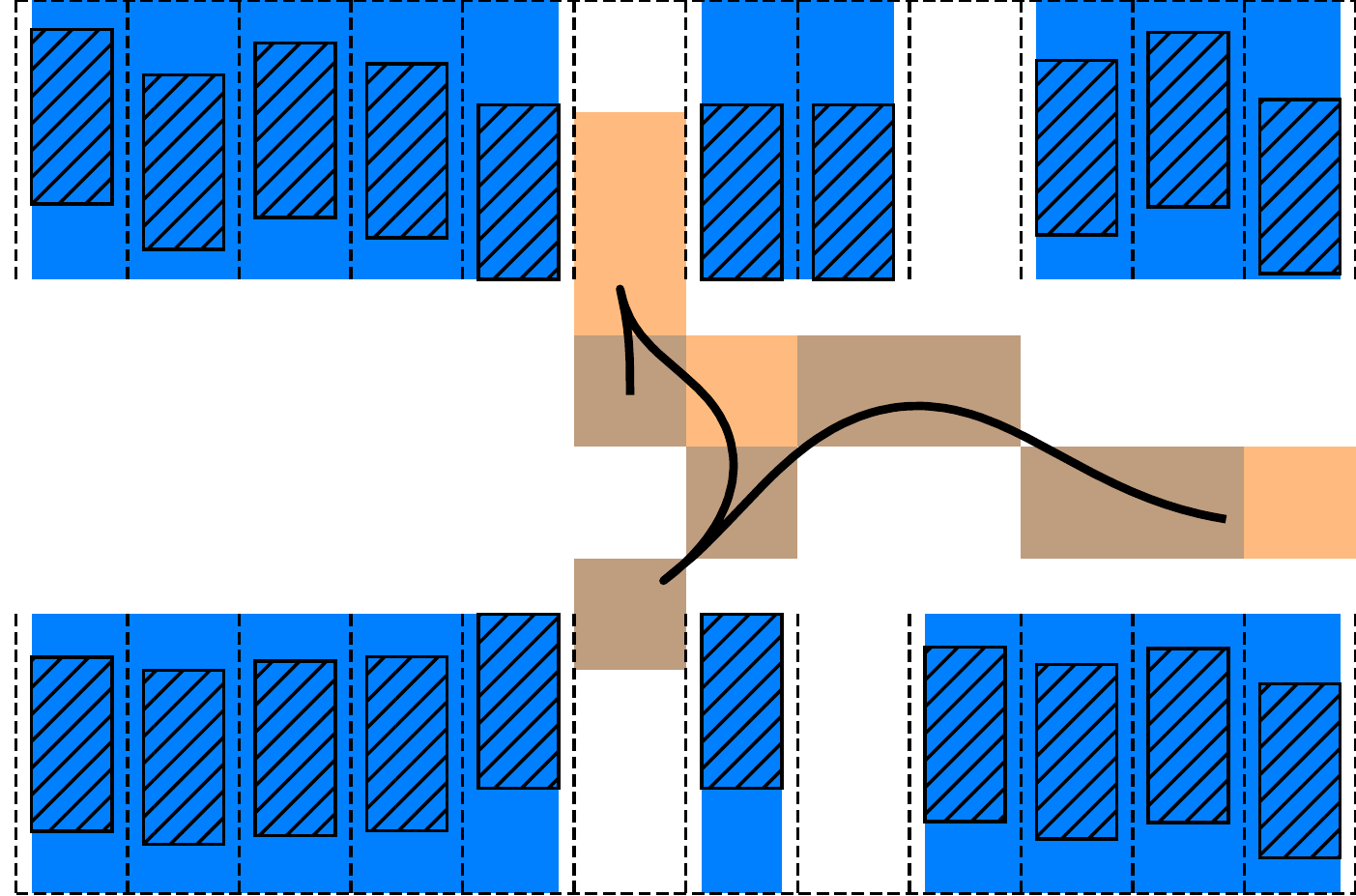}
		\caption{Strategy-guided trajectory}
		\label{fig:traj-ws-v0}
	\end{subfigure}
        \\
        \begin{subfigure}[t]{0.49\columnwidth}
		\centering
		\includegraphics[width=\textwidth]{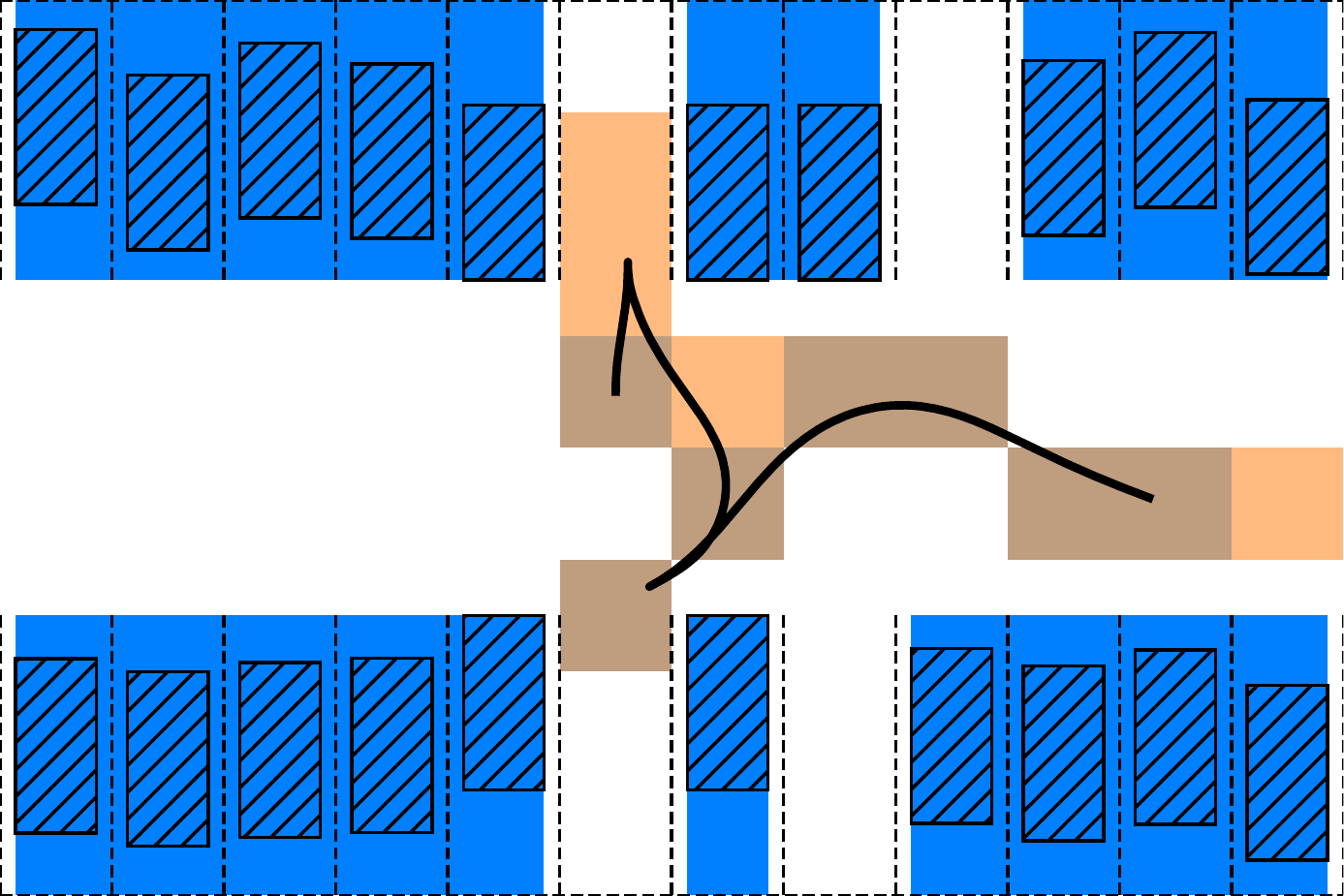}
	    \caption{Collision-free trajectory }
	    \label{fig:traj-final-v0}
	\end{subfigure}%
	~
	\begin{subfigure}[t]{0.49\columnwidth}
		\centering
		\includegraphics[width=\textwidth]{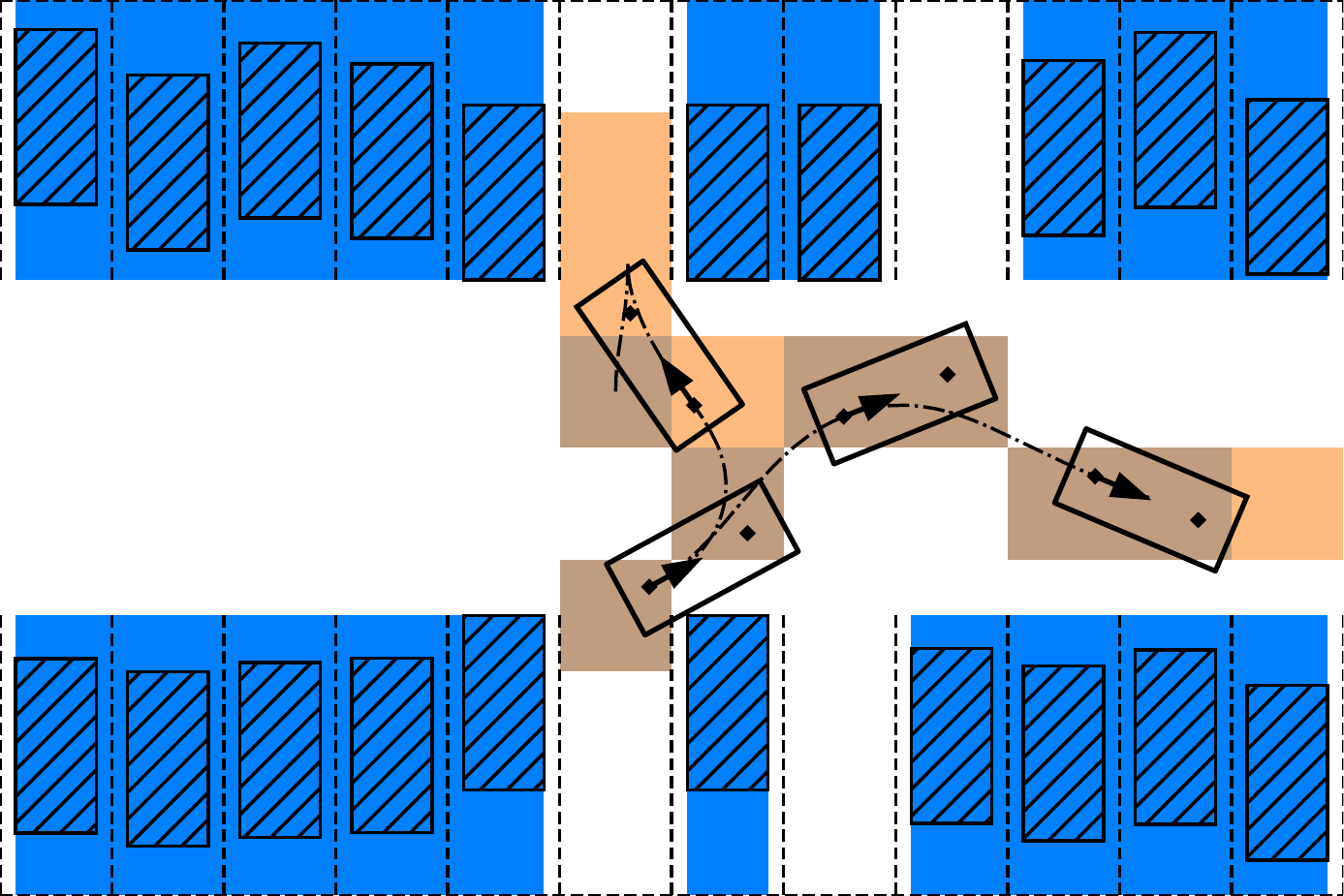}
		\caption{Poses along the trajectory}
		\label{fig:traj-final-pose-v0}
	\end{subfigure}
	\caption{The progressive steps of generating the single-vehicle trajectory for vehicle 0 under strategy-guided configuration constraints and collision avoidance constraints. There are overlaps among strategy-guided configuration sets at different time steps.}
    \label{fig:single-vehicle-planning}
\end{figure}

Given the strageties $\mathbf{s}^{[i]}$, we will solve the optimal control problem~\eqref{eq:complete-formulation}. The cost function is formulated as $c(z, u) = \psi^2 + v^2 w^2 + a^2 + 1$ to reflect the passenger comfort, the amount of actuation, and the time consumption.
The initial guess of the sampling time is $\Bar{T}_{\mathrm{s}}=3 \mathrm{s}$, and the minimal safety distance is $d_{\mathrm{min}} = 0.05 \mathrm{m}$. The full discretization of continuous problems is realized with orthogonal collocation on finite elements,~\cite{biegler_nonlinear_2010}. 5-th order Lagrange interpolation polynomial and Gauss-Radau roots are adopted for collocation. The optimization problems are implemented in CasADi,~\cite{andersson_casadi_2019}, and then solved by IPOPT,~\cite{wachter_implementation_2006}, with the linear solver HSL\_MA97,~\cite{rees_hsl_2022}.

Fig.~\ref{fig:single-vehicle-planning} illustrates the hierarchy of obtaining single-vehicle collision-free trajectories, as described in Section~\ref{sec:hierarchical-steps}:
\begin{enumerate}[label=(\roman*)]
    \item \textit{Pose Interpolation}: The piecewise Bézier curve is computed as Fig.~\ref{fig:spline-v0} for pose interpolation. Although the curve is smooth, it is neither kinematically feasible nor collision-free in the tight space.
    \item \textit{Strategy-guided Trajectory}: By using the Bézier curve as initial-guess, the single-vehicle strategy-guided trajectory $\Tilde{\mathbf{z}}^{[i]}$ is computed as Fig.~\ref{fig:traj-ws-v0}. The trajectory is kinematically feasible and compliant with the strategy-guided configurations.
    \item \textit{Collision-free Trajectory}: After adding the collision avoidance constraints, the trajectory $\Bar{\mathbf{z}}^{[i]}$ is optimized as Fig.~\ref{fig:traj-final-v0}, and the vehicle poses at different time steps are demonstrated in Fig.~\ref{fig:traj-final-pose-v0}.
\end{enumerate}

\begin{figure}
\begin{center}
\includegraphics[width=0.97\linewidth]{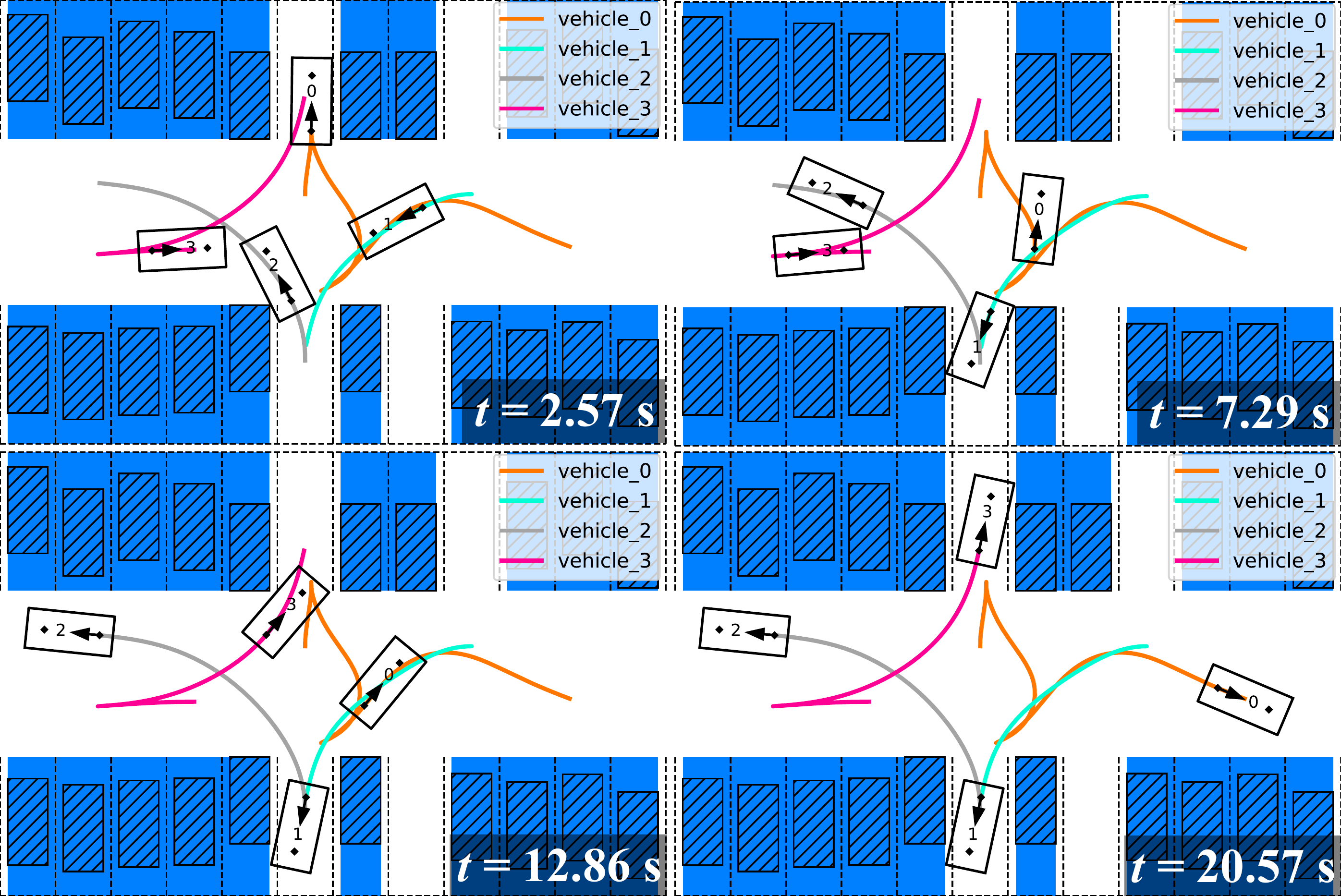}    
\caption{Final trajectories for successful conflict resolution. Snapshots are taken from four different time steps to illustrate the vehicle configurations.}
\label{fig:multi-vehicle-poses}
\end{center}
\end{figure}

\begin{figure}
\begin{center}
\includegraphics[width=\linewidth]{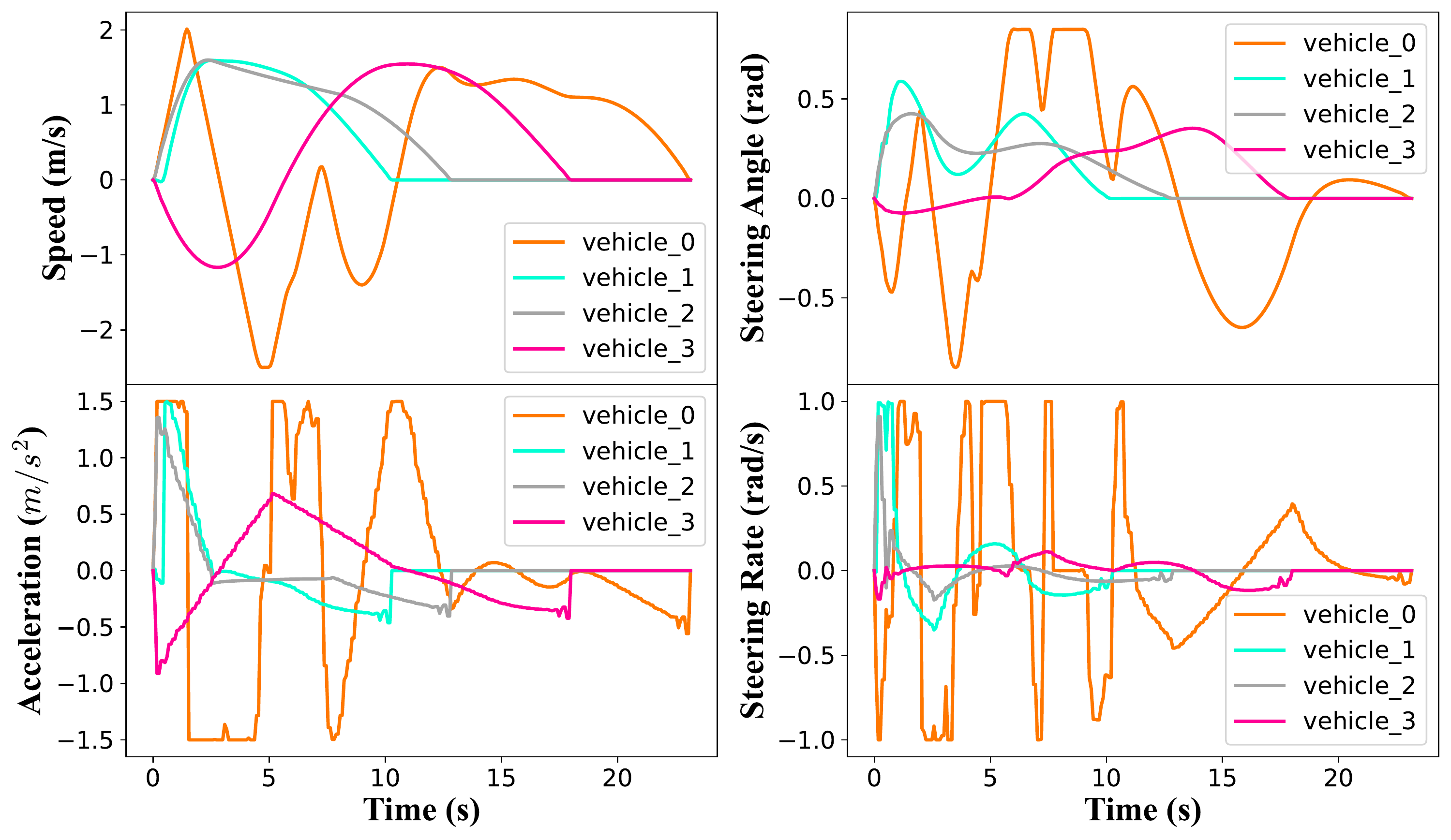}    
\caption{Profiles of vehicle speed $v$, front steering angle $\delta_f$, acceleration $a$, and steering rate $w$.}
\label{fig:multi-vehicle-states-inputs}
\end{center}
\end{figure}

Finally, the continuous trajectories for the multi-vehicle conflict resolution problem are obtained as Fig.~\ref{fig:multi-vehicle-poses}. The vehicle configurations are guided by strategies $\mathbf{s}^{[i]}$ such that vehicle 0 and vehicle 3 firstly make compromises to create more space, and vehicles 1 and 2 directly drive to their destinations. It can be observed from the snapshots that the vehicles can make dexterous collision-free maneuvers in tight spaces without any geometrical approximation. According to Fig.~\ref{fig:multi-vehicle-states-inputs}, vehicles obtain smooth speed and steering profiles, and all states and inputs are constrained within the operating limits.

\section{Conclusion}
\label{sec:conclusion}

This paper proposes a novel method to resolve multi-vehicle conflict in tightly-constrained spaces, which merges optimal control with reinforcement learning.
By starting from a simplified environment and guiding the optimal control problem with learned strategies, we can successfully approach the solution that is intractable to obtain from the initial formulation.

Firstly, the conflict resolution problem is transformed as a multi-agent POMDP in a discrete environment, so that we can leverage the off-the-shelf DQN algorithm to explore efficient actions for vehicles to reach their destinations while keeping collision-free. The trained policy generates strategies to guide tactical vehicle configurations in the optimal control problem. 
Then, since the vehicle dynamics and collision avoidance constraints are highly nonlinear, we provide a hierarchy to obtain high-quality initial guesses by progressively refining with simple NLP problems.

Future directions include exploring more powerful multi-agent RL algorithms so that the trained policy can be generalized to arbitrary environments, and decentralizing the optimal control formulation to reduce the computation burden for the conflict resolution problem at scale.


\bibliography{references}             

\begin{thebibliography}{18}
\providecommand{\natexlab}[1]{#1}
\providecommand{\url}[1]{\texttt{#1}}
\providecommand{\urlprefix}{URL }
\expandafter\ifx\csname urlstyle\endcsname\relax
  \providecommand{\doi}[1]{doi:\discretionary{}{}{}#1}\else
  \providecommand{\doi}{doi:\discretionary{}{}{}\begingroup
  \urlstyle{rm}\Url}\fi

\bibitem[{Andersson et~al.(2019)Andersson, Gillis, Horn, Rawlings, and
  Diehl}]{andersson_casadi_2019}
Andersson, J.A.E., Gillis, J., Horn, G., Rawlings, J.B., and Diehl, M. (2019).
\newblock {CasADi}: a software framework for nonlinear optimization and optimal
  control.
\newblock \emph{Mathematical Programming Computation}, 11(1), 1--36.
\newblock \doi{10.1007/s12532-018-0139-4}.

\bibitem[{Biegler(2010)}]{biegler_nonlinear_2010}
Biegler, L.T. (2010).
\newblock \emph{Nonlinear {Programming}: {Concepts}, {Algorithms}, and
  {Applications} to {Chemical} {Processes}}.
\newblock Society for Industrial and Applied Mathematics, USA.

\bibitem[{Campos et~al.(2014)Campos, Falcone, Wymeersch, Hult, and
  Sjöberg}]{campos_cooperative_2014}
Campos, G.R., Falcone, P., Wymeersch, H., Hult, R., and Sjöberg, J. (2014).
\newblock Cooperative receding horizon conflict resolution at traffic
  intersections.
\newblock In \emph{53rd {IEEE} {Conference} on {Decision} and {Control}},
  2932--2937.
\newblock \doi{10.1109/CDC.2014.7039840}.
\newblock ISSN: 0191-2216.

\bibitem[{Gupta et~al.(2017)Gupta, Egorov, and
  Kochenderfer}]{gupta_cooperative_2017}
Gupta, J.K., Egorov, M., and Kochenderfer, M. (2017).
\newblock Cooperative {Multi}-agent {Control} {Using} {Deep} {Reinforcement}
  {Learning}.
\newblock In G.~Sukthankar and J.A. Rodriguez-Aguilar (eds.), \emph{Autonomous
  {Agents} and {Multiagent} {Systems}}, Lecture {Notes} in {Computer}
  {Science}, 66--83. Springer International Publishing, Cham.
\newblock \doi{10.1007/978-3-319-71682-4-5}.

\bibitem[{Katriniok et~al.(2017)Katriniok, Kleibaum, and
  Joševski}]{katriniok_distributed_2017}
Katriniok, A., Kleibaum, P., and Joševski, M. (2017).
\newblock Distributed {Model} {Predictive} {Control} for {Intersection}
  {Automation} {Using} a {Parallelized} {Optimization} {Approach}.
\newblock \emph{IFAC-PapersOnLine}, 50(1), 5940--5946.
\newblock \doi{10.1016/j.ifacol.2017.08.1492}.

\bibitem[{Li et~al.(2018)Li, Kolmanovsky, Girard, and Yildiz}]{li_game_2018}
Li, N., Kolmanovsky, I., Girard, A., and Yildiz, Y. (2018).
\newblock Game {Theoretic} {Modeling} of {Vehicle} {Interactions} at
  {Unsignalized} {Intersections} and {Application} to {Autonomous} {Vehicle}
  {Control}.
\newblock In \emph{2018 {Annual} {American} {Control} {Conference} ({ACC})},
  3215--3220.
\newblock \doi{10.23919/ACC.2018.8430842}.
\newblock ISSN: 2378-5861.

\bibitem[{Li et~al.(2019)Li, Egorov, and Kochenderfer}]{li_optimizing_2019}
Li, S., Egorov, M., and Kochenderfer, M. (2019).
\newblock Optimizing {Collision} {Avoidance} in {Dense} {Airspace} using {Deep}
  {Reinforcement} {Learning}.
\newblock \doi{10.48550/arXiv.1912.10146}.

\bibitem[{Murgovski et~al.(2015)Murgovski, de~Campos, and
  Sjöberg}]{murgovski_convex_2015}
Murgovski, N., de~Campos, G.R., and Sjöberg, J. (2015).
\newblock Convex modeling of conflict resolution at traffic intersections.
\newblock In \emph{2015 54th {IEEE} {Conference} on {Decision} and {Control}
  ({CDC})}, 4708--4713.
\newblock \doi{10.1109/CDC.2015.7402953}.

\bibitem[{Raﬃn et~al.(2021)Raﬃn, Hill, Gleave, Kanervisto, Ernestus, and
  Dormann}]{ran_stable-baselines3_2021}
Raﬃn, A., Hill, A., Gleave, A., Kanervisto, A., Ernestus, M., and Dormann, N.
  (2021).
\newblock Stable-{Baselines3}: {Reliable} {Reinforcement} {Learning}
  {Implementations}.
\newblock \emph{Journal of Machine Learning Research}, 22(268), 8.

\bibitem[{Rees(2022)}]{rees_hsl_2022}
Rees, T. (2022).
\newblock {HSL}. {A} collection of {Fortran} codes for large scale scientific
  computation.
\newblock \urlprefix\url{http://www.hsl.rl.ac.uk/}.

\bibitem[{Rey et~al.(2018)Rey, Pan, Hauswirth, and Lygeros}]{rey_fully_2018}
Rey, F., Pan, Z., Hauswirth, A., and Lygeros, J. (2018).
\newblock Fully {Decentralized} {ADMM} for {Coordination} and {Collision}
  {Avoidance}.
\newblock In \emph{2018 {European} {Control} {Conference} ({ECC})}, 825--830.
\newblock \doi{10.23919/ECC.2018.8550245}.

\bibitem[{Riegger et~al.(2016)Riegger, Carlander, Lidander, Murgovski, and
  Sjöberg}]{riegger_centralized_2016}
Riegger, L., Carlander, M., Lidander, N., Murgovski, N., and Sjöberg, J.
  (2016).
\newblock Centralized {MPC} for autonomous intersection crossing.
\newblock In \emph{2016 {IEEE} 19th {International} {Conference} on
  {Intelligent} {Transportation} {Systems} ({ITSC})}, 1372--1377.
\newblock \doi{10.1109/ITSC.2016.7795736}.
\newblock ISSN: 2153-0017.

\bibitem[{Terry et~al.(2021)Terry, Black, Grammel, Jayakumar, Hari, Sullivan,
  Santos, Perez, Horsch, Dieffendahl, Williams, Lokesh, and
  Ravi}]{terry_pettingzoo_2021}
Terry, J.K., Black, B., Grammel, N., Jayakumar, M., Hari, A., Sullivan, R.,
  Santos, L., Perez, R., Horsch, C., Dieffendahl, C., Williams, N.L., Lokesh,
  Y., and Ravi, P. (2021).
\newblock {PettingZoo}: {Gym} for {Multi}-{Agent} {Reinforcement} {Learning}.
\newblock \doi{10.48550/arXiv.2009.14471}.
\newblock ArXiv:2009.14471 [cs, stat].

\bibitem[{Terry et~al.(2022)Terry, Grammel, Son, and
  Black}]{terry_parameter_2022}
Terry, J.K., Grammel, N., Son, S., and Black, B. (2022).
\newblock Parameter {Sharing} {For} {Heterogeneous} {Agents} in {Multi}-{Agent}
  {Reinforcement} {Learning}.
\newblock ArXiv:2005.13625 [cs, stat].

\bibitem[{Wächter and Biegler(2006)}]{wachter_implementation_2006}
Wächter, A. and Biegler, L.T. (2006).
\newblock On the implementation of an interior-point filter line-search
  algorithm for large-scale nonlinear programming.
\newblock \emph{Mathematical Programming}, 106(1), 25--57.
\newblock \doi{10.1007/s10107-004-0559-y}.

\bibitem[{Yuan et~al.(2022)Yuan, Shan, and Mi}]{yuan_deep_2022}
Yuan, M., Shan, J., and Mi, K. (2022).
\newblock Deep {Reinforcement} {Learning} {Based} {Game}-{Theoretic}
  {Decision}-{Making} for {Autonomous} {Vehicles}.
\newblock \emph{IEEE Robotics and Automation Letters}, 7(2), 818--825.
\newblock \doi{10.1109/LRA.2021.3134249}.
\newblock Conference Name: IEEE Robotics and Automation Letters.

\bibitem[{Zhang et~al.(2020)Zhang, Liniger, and
  Borrelli}]{zhang_optimization-based_2020}
Zhang, X., Liniger, A., and Borrelli, F. (2020).
\newblock Optimization-{Based} {Collision} {Avoidance}.
\newblock \emph{IEEE Transactions on Control Systems Technology}, 1--12.
\newblock \doi{10.1109/TCST.2019.2949540}.
\newblock ArXiv: 1711.03449.

\bibitem[{Zhang et~al.(2019)Zhang, Liniger, Sakai, and
  Borrelli}]{zhang_autonomous_2019}
Zhang, X., Liniger, A., Sakai, A., and Borrelli, F. (2019).
\newblock Autonomous {Parking} {Using} {Optimization}-{Based} {Collision}
  {Avoidance}.
\newblock In \emph{Proceedings of the {IEEE} {Conference} on {Decision} and
  {Control}}, volume 2018-Decem, 4327--4332. IEEE.
\newblock \doi{10.1109/CDC.2018.8619433}.
\newblock ISSN: 07431546.

\end{thebibliography}
                                                   







\end{document}